\documentclass[conference]{IEEEtran}
\IEEEoverridecommandlockouts

\usepackage{times}
\usepackage{epsfig}
\usepackage{graphicx}
\usepackage{amsmath}
\usepackage{amssymb}
\usepackage{url}
\usepackage{algorithmic}
\usepackage{algorithm}

\usepackage{booktabs} 
\usepackage{enumerate}
\usepackage{adjustbox}
\usepackage{indentfirst}
\usepackage{array}
\usepackage{multirow}
\usepackage{float}
\usepackage{hhline}
\usepackage{bm}
\usepackage{subfigure}

\usepackage{fancyhdr}
\fancypagestyle{firstpage}{%
  \rhead{IEEE CICT 2019}
  \lfoot{\small{\copyright2019 IEEE. 
Personal use of this material is permitted. Permission from IEEE must be obtained for all other users, including reprinting/republishing this material for advertising or promotional purposes, creating new collective works for resale or redistribution to servers or lists, or reuse of any copyrighted components of this work in other works.}}
}

\begin{document}

\title{PSNet: Parametric Sigmoid Norm Based CNN for Face Recognition}

\author{Yash Srivastava, Vaishnav Murali and Shiv Ram Dubey\\
Computer Vision Group,\\
Indian Institute of Information Technology, Sri City, Chittoor, Andhra Pradesh, India\\
\{srivastava.y15, murali.v15, srdubey\}@iiits.in}

\maketitle
\thispagestyle{firstpage}

\begin{abstract}
The Convolutional Neural Networks (CNN) have become very popular recently due to its outstanding performance in various computer vision applications. It is also used over widely studied face recognition problem. However, the existing layers of CNN are unable to cope with the problem of hard examples which generally produce lower class scores. Thus, the existing methods become biased towards the easy examples. In this paper, we resolve this problem by incorporating a Parametric Sigmoid Norm (PSN) layer just before the final fully-connected layer. We propose a PSNet CNN model by using the PSN layer. The PSN layer facilitates high gradient flow for harder examples as compared to easy examples. Thus, it forces the network to learn the visual characteristics of hard examples. We conduct the face recognition experiments to test the performance of PSN layer. The suitability of the PSN layer with different loss functions is also experimented. The widely used Labeled Faces in the Wild (LFW) and YouTube Faces (YTF) datasets are used in the experiments. The experimental results confirm the relevance of the proposed PSN layer.
\end{abstract}

\begin{IEEEkeywords}
Face Recognition, Deep Learning, Sigmoid Function, Parametric Sigmoid Layer, PSNet.
\end{IEEEkeywords}

\section{Introduction}
The face recognition is a very challenging problem due to several factors ranging from the personal to environmental factors. A person can make different appearances of the face by changing the hair style, beard style, cap, glasses, etc. Many times the capturing device is not able to capture very good face image and contains various challenges such as difficult pose, bad lighting condition, etc. However, the face based recognition system has a huge potential to automate many applications in real-life. 

\begin{figure*}[!t]
\centering
\includegraphics[width=\linewidth]{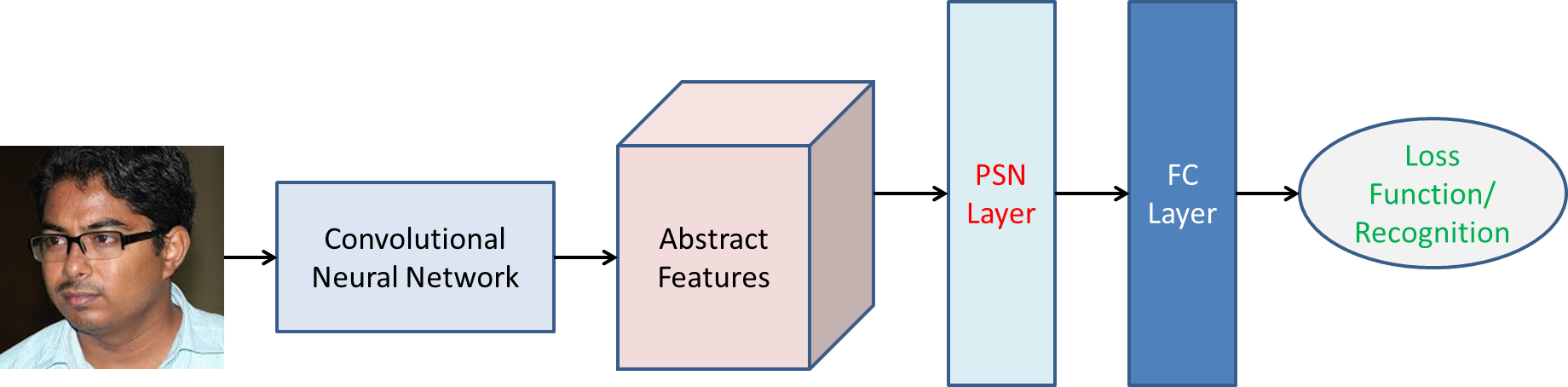}
\caption{The proposed PSNet CNN model for face recognition with Parametric-Sigmoid-Norm (PSN) layer represented. Given an input image, the image is passed through the convolutional layers of the CNN and features are obtained just before the final class score layer. These feature tensors are passed through the Parametrize-Sigmoid-Norm layer to enforce the feature distribution as close as possible for both easy as well as hard examples. It leads to better learning of hard examples also along with easy examples. The face image taken in this illustration is provided by one of the author this paper.}
\label{fig:workflow}
\end{figure*}

Nowadays, the deep learning has pushed the performance of any recognition system with a great success in forward direction. The first Convolutional Neural Network (CNN), named as AlexNet \cite{krizhevsky2012imagenet} introduced by Krizhevsky et al. completely changed the dynamics in visual recognition area. Finally, the AlexNet became a revolutionary model for the task of image classification. It also won the ImageNet Large Scale Challenge in 2012. After that, various CNN based architectures have been introduced for different applications such as VGGNet \cite{simonyan2014very}, GoogleNet \cite{szegedy2015going}, ResNet \cite{he2016deep} for image classification; Local Bit-plane Decoded CNN \cite{dubeylocal}, ChexNet \cite{rajpurkar2017chexnet} and RCCNet \cite{basha2018rccnet} for biomedial image retrieval/classification; CNN for face spoof detection \cite{nagpal2018performance}, \cite{liu2018learning}; and many more. 

The CNN has also shown its impact in the face recognition area as well. The DeepFace \cite{taigman2014deepface}, DeepID2 \cite{sun2014deep}, FaceNet \cite{schroff2015facenet} and two-branch CNN \cite{zangeneh2020low} are few state-of-the-art methods which have shown a significant performance gain when compared to the hand-crafted features. The above growth is accompanied by the development of large-scale face datasets for training and testing of the CNN models. The major face datasets include CASIA-Webface \cite{yi2014learning}, Labeled Faces in the Wild (LFW) \cite{lfw} and Youtube Faces (YTF) \cite{ytf} among other face datasets.

The study of Convoultional Neural Networks over time have shown that the deep CNN architectures are generally better in performance when compared to the shallower networks. It leads to the rise of deep architectures like ResNet \cite{he2016deep} and Inception \cite{szegedy2017inception} which use multiple trainable convolutional layers to achieve higher accuracy. At the same time, experiments showed that after certain depth, not much increment in accuracy was observed, i.e., the performance saturated at great depths \cite{he2016deep}. Moreover, the rise of large-scale face recognition leading to heavy architectures would require significant resources to run the recognition systems which hinders its applications on resource constrained devices. Considering the above two points, researchers have started exploring and working over different parts of the architecture system like non-linearity layers, loss functions, and many more. The Rectified Linear Unit (ReLU) \cite{krizhevsky2012imagenet}, Parametric ReLU \cite{he2015delving}, Average Biased ReLU \cite{dubey2018average}, and Linearly Scaled Hyperbolic Tangent (LiSHT) \cite{roy2019lisht}. The work in loss functions has been quite significant with recent state-of-the-art functions like ArcFace \cite{deng2018arcface}, SphereFace \cite{liu2017sphereface}, and Hard-Mining Loss \cite{srivastava2019hard}, specifically designed for the face recognition task. Other type of layers used in CNN is Dropconnect \cite{wan2013regularization}, Dropout \cite{srivastava2014dropout}, Batch Normalization \cite{ioffe2015batch}, etc. A performance analysis of loss functions over face datasets is conducted in \cite{srivastava2019performance}.

Most of these types of layers are proposed to increase the learning capacity and generalization ability of CNN models. However, it turns out that after a certain limit the correctly classified samples introduce the biasedness and hampers the learning of incorrectly classified samples. We believe that it happens because the abstract features for easy examples become different from the hard examples. In this paper, we tackle this problem by introducing a parametric sigmoid norm layer just before the final class score layer to force the abstract features of both easy and hard examples to be closer to each other. It leads to the better training of model for harder examples also.

In this paper, first we present the proposed Parametric-Sigmoid-Norm (PSN) layer in Section 2. We also discuss its importance just before the final fully-connected layer. Next, we present the experimental setup and details about the architectures and datasets used in Section 3. The experimental results and analysis are described in Section 4. Finally, the concluding remarks are summarized with future directions in Section 5.

\begin{table*}[!t]
\caption{The experimental results with different settings of proposed PSNet model (i.e., the Parametric-Sigmoid-Norm (PSN) on ResNet architectures). The results are compared using PSNet18 (i.e., ResNet18 with PSN) and PSNet50 (i.e., ResNet50 with PSN) models. The training is done over CASIA-WebFace face dataset. However, the testing is performed using LFW and YTF face datasets. Here, the term $Trainable$ means the corresponding hyperparameter is learnt automatically through training.}
\centering
\begin{tabular}{|m{1.4cm}|m{7.1cm}|m{2.5cm}|m{2.5cm}|}
\hline
\textbf{Model} &  \textbf{Parametric-Sigmoid-Norm Hyperparameters} &  \textbf{Accuracy on LFW} & \textbf{Accuracy on YTF} \tabularnewline
\hline
\hline
\multirow{7}{*}{PSNet18} & $Without \; PSN \text{ (i.e., ResNet18)}$ & 95.35 & 91.80 \tabularnewline
\cline{2-4}
& $\alpha=1, \beta=20, \gamma=1$ & 95.68 & 92.60 \tabularnewline
\cline{2-4}
& $\alpha=Trainable, \beta=20, \gamma=1$ & 95.12 & 91.90 \tabularnewline
\cline{2-4}
& $\alpha=1, \beta=Trainable, \gamma=1$ & 96.40 & 93.34 \tabularnewline
\cline{2-4}
& $\alpha=1, \beta=20, \gamma=Trainable$ & 96.09 & 93.43 \tabularnewline
\cline{2-4}
& $\alpha=1, \beta=Trainable, \gamma=Trainable$ & \textbf{96.87} & \textbf{94.10} \tabularnewline
\cline{2-4}
& $\alpha=Trainable, \beta=Trainable,\gamma=Trainable$ & 95.95 & 93.20 \tabularnewline
\cline{1-4}\cline{1-4}\cline{1-4}\cline{1-4}
\multirow{7}{*}{PSNet50} & $Without \; PSN \text{ (i.e., ResNet50)}$ & 97.42 & 94.50 \tabularnewline
\cline{2-4}
& $\alpha=1, \beta=20, \gamma=1$ & 97.50 & 94.47 \tabularnewline
\cline{2-4}
& $\alpha=Trainable, \beta=20, \gamma=1$ & 96.98 & 94.34 \tabularnewline
\cline{2-4}
& $\alpha=1, \beta=Trainable, \gamma=1$ & 97.71 & 94.64 \tabularnewline
\cline{2-4}
& $\alpha=1, \beta=20, \gamma=Trainable$ & 97.89 & 94.67 \tabularnewline
\cline{2-4}
& $\alpha=1, \beta=Trainable, \gamma=Trainable$ & \textbf{98.10} & \textbf{94.90} \tabularnewline
\cline{2-4}
& $\alpha=Trainable, \beta=Trainable, \gamma=Trainable$ & 97.53 & 94.33 \tabularnewline
\cline{1-4}\cline{1-4}\cline{1-4}
\hline
\end{tabular}
\label{table:results_comparison}
\end{table*}

\begin{table*}[!t]
\caption{The experimental results comparison using Angular-Softmax and ArcFace loss functions over ResNet and PSNet CNN models. The results are compared for different network depth such as 18 and 50. The training is performed over CASIA-WebFace dataset. The testing is performed using LFW and YTF face datasets. The hyperparameter settings in PSNet are kept as $\alpha=1$ and $\beta$ \& $\gamma$ are network trainable.}
\centering
\begin{tabular}{|c|c|c|c|}
\cline{1-4}\cline{1-4}\cline{1-4}\cline{1-4}\cline{1-4}
\textbf{Loss Function} & \textbf{Architecture} &  \textbf{Accuracy on LFW} & \textbf{Accuracy on YTF} \tabularnewline
\cline{1-4}\cline{1-4}\cline{1-4}\cline{1-4}\cline{1-4}
\multirow{2}{*}{Angular-Softmax} & ResNet18 & 97.12 & 93.90 \tabularnewline
\cline{2-4}
& PSNet18 & \textbf{97.28} & \textbf{94.22} \tabularnewline
\cline{1-4}\cline{1-4}\cline{1-4}\cline{1-4}\cline{1-4}

\multirow{2}{*}{ArcFace} & ResNet18 & \textbf{97.79} & 94.54 \tabularnewline
\cline{2-4}
& PSNet18 & 97.78 & \textbf{94.65} \tabularnewline
\cline{1-4}\cline{1-4}\cline{1-4}\cline{1-4}\cline{1-4}

\multirow{2}{*}{Angular-Softmax} & ResNet50 & 99.10 & 96.10 \tabularnewline
\cline{2-4}
& PSNet50 & \textbf{99.21} & \textbf{96.50} \tabularnewline
\cline{1-4}\cline{1-4}\cline{1-4}\cline{1-4}\cline{1-4}

\multirow{2}{*}{ArcFace} & ResNet50 & 99.20 & 96.30 \tabularnewline
\cline{2-4}
& PSNet50 & \textbf{99.26} & \textbf{97.10} \tabularnewline
\cline{1-4}\cline{1-4}\cline{1-4}\cline{1-4}\cline{1-4}
\end{tabular}
\label{table:results_loss}
\end{table*}

\section{Proposed Parametric-Sigmoid-Norm Layer}
Considering a range of available face datasets like CASIA-Webface \cite{yi2014learning}, LFW \cite{lfw} among others, we observe that these datasets contain a large amount of face images belonging to a single identity. Due to this, there is be a big cluster of easy samples which might be far in feature space from the significantly small cluster of hard samples when the distance is measured between the centroids of both these clusters. This leads to a bias in the model which is trained on such a dataset and might perform poorly in real-case scenarios. We tackle this problem by enforcing the features to follow the parametric sigmoid norm. Generally, sigmoid is used as the loss as it converts the input into the range of 0 and 1. We leverage this property of sigmoid over the CNN features before class score computation. The sigmoid is defined as follows:
\begin{equation}
\mathcal{F}_{{Sigmoid}}(x)= \frac{1}{1 + e^{-x}}
\label{eq:sigmoid1}
\end{equation}
where $x$ is the input and $\mathcal{F}_{{Sigmoid}}(x) \in [0,1]$. Note that using this sigmoid over abstract CNN features might disturb the distribution in an adverse manner. In order to avoid it, we use the parametric sigmoid which allows some flexibility in sigmoid using some parameters. The parametric sigmoid function is defined as follows: 
\begin{equation}
\mathcal{F}_{{PSigmoid}}(x)= \frac{\alpha}{1 + e^{-\beta(x-\gamma)}}
\label{eq:sigmoid2}
\end{equation}
where $\alpha$, $\beta$ and $\gamma$ are the parameters and/or hyper-parameters which have been kept either trainable or fixed under different setting scenarios. Keeping $\alpha$ equal to $1$, the parametric-sigmoid function clips the output value between $0$ to $1$. Note that the proposed PSN layer uses $\mathcal{F}_{{PSigmoid}}$ function for all its inputs. The effect of the application of the above function helps to bring the centroids of hard and easy sample clusters closer in abstract feature space. Hence, making it easier for the model to learn the training dataset irrespective of easy or hard examples.

We use the parametric-sigmoid-norm (PSN) layer with the existing CNN as shown in Fig. \ref{fig:workflow}. The PSN layer is used just before the final fully-connected layer (i.e., class score layer) to enforce the abstract features of all samples to follow a certain distribution irrespective of its difficulty level. Basically, it tries to reduce the cluster distance between easy and hard samples for a class. In other words, it generalizes the capacity of model with high priority for harder examples.

\section{Experimental Setup}
In this section, first we introduce the CNN models used for face recognition, then we provide a summary of training and testing datasets used, and finally we state the other settings related to training of networks.

\subsection{CNN Architectures}
The residual networks (ResNet) have shown very promising performance in various applications of computer vision \cite{he2016deep}. We use the ResNet model along with proposed parametric sigmoid norm (PSN) layer as the PSNet model. Thus, PSNet18 model consists of ResNet18 and PSN layer and PSNet50 model consists of ResNet50 and PSN layer.

\subsection{Training Datasets}
The CASIA-Webface face dataset \cite{yi2014learning} is adapted in this paper for the training purpose. This dataset is used for training in most of the recent work of the face recognition. This dataset has 10,575 persons leading to total of 4,94,414 face images. Thus, the number of classes is 10,575 in CASIA-WebFace dataset.

\subsection{Testing Datasets}
The Labeled Faces in the Wild (LFW) \cite{lfw} and YouTube Faces (YTF) \cite{ytf} are used as our testing datasets. A total of 5749 different people leading to 13233 images is present in the LFW dataset. However, 3425 videos from 1595 identities are present in the YTF dataset. The image frame numbers are provided to be used for training and testing purpose in the YTF dataset. We use the standard as per the LFW benchmark for face verification and provide the performance over both testing datasets in terms of the accuracy. These accuracies have been used as the performance measure in face recognition works and hence reported here also for comparison.

\subsection{Input Data and Network Settings}
We use the MTCNN \cite{zhang2016joint} model for the alignment of the face images similar to \cite{deng2018arcface}, \cite{liu2017sphereface}. A normalization of face images is performed by subtracting 127.5 from each pixel and then divided by 128. The batch-size is kept at 64 with . The training is performed for 20 epochs for each model. The learning rate is initialized with 0.01 and reduced by a factor of 10 at the 8th, 12th and 16th epochs. The Stochastic Gradient Descent (SGD) with momentum optimization technique is used with default settings for training using backpropagation.

\section{Experimental Results and Observations}
In this section, first we present the experimental results analysis of the proposed method and then we show its generalization using different loss functions. 

\subsection{Result Analysis}
We summarize the experimental results in Table 1 using the proposed PSNet model with the PSN layer with depth 18 (i.e., PSNet18) and 50 (i.e., PSNet50). The training is done over CASIA-WebFace dataset and testing is done over LFW and YTF datasets. The Cross-entropy loss is used in this experiment. The results are compared under different settings of PSN layer such as $\alpha$, $\beta$, and $\gamma$ are either fixed or trainable. If it is not trainable, then it is fixed based empirical observation. Note that under first setting, PSN layer is not used, thus it is same as the ResNet model. It is discovered that the performance of PSNet18 as well as PSNet50 is better when $\alpha=1$ and $\beta$ and $\gamma$ are trained using backpropagation. The performance of PSNet18 and PSNet50 is significantly improved as compared to ResNet18 and ResNet50, respectively under best setting of PSN layer. Another important finding is that the proposed PSNet performs relatively much better when the depth of the network is less which can lead to better lightweight models for resource constrained devices.

\subsection{PSNet with Other Losses}
In this paper, we compare the results with existing methods in terms of the existing loss functions. The Cross-entropy loss is used to compute the results discussed in the previous section. In order to show the generalization ability of the proposed PSNet model, we also compute the results of ResNet and PSNet using specialized loss functions such as Angular-Softmax \cite{liu2017sphereface} and ArcFace \cite{deng2018arcface}. Table 2 illustrates the results for ResNet18, PSNet18, ResNet50, and PSNet50 using Angular-Softmax and ArcFace losses. The training is done using CASIA-WebFace dataset and testing is done over LFW and YTF face datasets. It is evident from the experimental results using different loss functions that the proposed PSNet model outperforms the state-of-the-art ResNet model for the same depth. It shows the importance of PSN layer which converts some of hard examples into easy by enforcing the parametric sigmoid norm in abstract feature space.

As we can see from the experimental results, with Loss functions like ArcFace and SphereFace, there is an improvement in performance with the usage of shallower networks. The proposed PSN layer is added to reduce the distance between hard and easy faces. The added robustness leads to the improvement in accuracy over LFW and YTF faces.

\section{Conclusion}
This paper introduces a parametric sigmoid norm (PSN) layer in the abstract feature space of CNN to enforce the better learning of data irrespective of its difficulty level. The proposed PSNet model introduces the PSN layer just before the final class score layer of ResNet. The face recognition experiments are performed to judge the performance of the proposed model. The results are computed over widely adapted LFW and YTF face datasets while the training is performed over CASIA-WebFace dataset. It turns out from the experiments that the hyperparameters $\alpha=1$ while $\beta$ and $\gamma$ as trainable lead to better performance of PSNet. The proposed model is also tested with state-of-the-art loss functions such as Angular-Softmax and ArcFace and has shown the improved performance. It is also observed that the proposed idea has a great impact over the relatively shallow model. It can lead to a better light weight CNN model for resource constrained devices.

\section*{Acknowledgment}
This research is funded by Science and Engineering Research Board (SERB), Govt. of India under Early Career Research (ECR) scheme through SERB/ECR/2017/000082 project fund.

\bibliographystyle{IEEEtran}
\bibliography{references}

\end{document}